\pdfoutput=1

\documentclass[11pt]{article}

\usepackage[final]{ACL2023}
\usepackage{flushend}

\usepackage{times}
\usepackage{latexsym}

\usepackage[T1]{fontenc}

\usepackage[utf8]{inputenc}

\usepackage{microtype}

\usepackage{inconsolata}

\usepackage{enumitem}
\usepackage{microtype}
\usepackage{graphicx}

\usepackage{subfigure}
\usepackage{booktabs} %
\usepackage{times}
\usepackage{latexsym}
\usepackage{multirow}
\usepackage{color}
\usepackage{siunitx}
\usepackage{algorithmic, algorithm}
\usepackage{setspace}
\usepackage{multirow}
\usepackage{multicol}

\usepackage{titlesec}
\usepackage{boldline}
\usepackage{pifont}%
\usepackage{amsmath,amstext, amsthm}
\usepackage{tablefootnote}
\definecolor{ForestGreen}{RGB}{34,139,34}
\graphicspath{{figures/}{../figures/}}

\usepackage{titlesec}

\usepackage{etoolbox}

\usepackage{caption}

\newcommand{\sref}[1]{Section~\ref{#1}}  \newcommand{\eref}[1]{Equation~(\ref{#1})} 
\newcommand{\tref}[1]{Table~\ref{#1}}

\def\x{{\mathbf x}} 
 
\def\y{{\mathbf y}} 
\def\X{{\mathbf X}} 
 
\def\Y{{\mathbf Y}}

\def\A{{\mathbf A}} 
\def\E{{\mathbf E}} 
 
\def\L{{\mathbf L}}

\def\F{{\mathcal F}} 
\def\X{{\mathcal X}} 
\def\Y{{\mathcal Y}}

\def\T{{\mathcal T}} 
\def\G{{\mathcal G}}

\newcommand{\tae}[0]{\texttt{ToxicTrap} }

\newcommand{\taee}[0]{\texttt{ToxicTraps Extend} }

\newcommand{\eat}[0]{\texttt{AT2} }
\newcommand{\method}[0]{\texttt{AT2} }

\newcommand{\sat}[0]{\texttt{AT1} }
\newcommand{\satdel}[0]{\texttt{AT1-delete} }
\newcommand{\satunk}[0]{\texttt{AT1-unk} }

\newcommand{\sdel}[0]{\texttt{delete} }
\newcommand{\sunk}[0]{\texttt{unk} }
\newcommand{\sws}[0]{\texttt{wt-saliency} }
\newcommand{\sgen}[0]{\texttt{genetic} }
\newcommand{\sgrad}[0]{\texttt{gradient} }
\newcommand{\sbeam}[0]{\texttt{beam} }

\newcommand{\tglove}[0]{\texttt{glove} }
\newcommand{\twn}[0]{\texttt{wordnet} }
\newcommand{\tmlm}[0]{\texttt{mlm} }

\newcommand{\noat}[0]{No \texttt{AT} }

\newcommand{\at}[0]{\texttt{AT} }
\newcommand{\jigsawb}[0]{\texttt{Jigsaw-BL} }
\newcommand{\jigsawm}[0]{\texttt{Jigsaw-ML} }
\newcommand{\tweetm}[0]{\texttt{HTweet-MC} }

\newcommand{\Jigsawb}[0]{\texttt{Jigsaw-BL} }
\newcommand{\Jigsawm}[0]{\texttt{Jigsaw-ML} }

\newcommand{\atot}[0]{\texttt{TT-A2T} }
\newcommand{\deepbug}[0]{\texttt{TT-Dbug} }
\newcommand{\textbugger}[0]{\texttt{TT-TBug} }
\newcommand{\textfooler}[0]{\texttt{TT-TFool} }
\newcommand{\pwws}[0]{\texttt{TT-Pwws} }

\newcommand{\atotO}[0]{\texttt{A2T} }
\newcommand{\deepbugO}[0]{\texttt{DeepWordBug} }
\newcommand{\textbuggerO}[0]{\texttt{TextBugger} }
\newcommand{\textfoolerO}[0]{\texttt{TextFooler} }
\newcommand{\pwwsO}[0]{\texttt{PWWS} }

\DeclareMathOperator*{\argmin}{arg\,min}

\usepackage{colortbl}

\usepackage[T2A]{fontenc}

\title{Towards Building a Robust Toxicity Predictor}

\author{ Dmitriy Bespalov, Sourav Bhabesh, Yi Xiang, Liutong Zhou, Yanjun Qi \\
  Amazon Web Services Bedrock Science \\
  \texttt{ \{ dbespal, sbhabesh, yxxan, yanjunqi \}@amazon.com} \\}

\begin{document}
\maketitle

\begin{abstract}

Recent NLP literature pays little attention to the robustness of toxicity language predictors, while these systems are most likely to be used in adversarial contexts. This paper presents a novel adversarial attack, \texttt{ToxicTrap}, introducing small word-level perturbations to fool SOTA text classifiers to predict toxic text samples as benign. \texttt{ToxicTrap} exploits greedy based search strategies to enable fast and effective generation of toxic adversarial examples. Two novel goal function designs allow \texttt{ToxicTrap} to identify weaknesses in both multiclass and multilabel toxic language detectors. Our empirical results show that SOTA toxicity text classifiers are indeed vulnerable to the proposed attacks, attaining over 98\% attack success rates in multilabel cases. We also show how a vanilla adversarial training and its improved version can help increase robustness of a toxicity detector even against unseen attacks.
\end{abstract}

\section{Introduction}

Deep learning-based natural language processing (NLP) plays a crucial role in detecting toxic language content~\citep{Ibrahim18,Zhao2019DetectingTC, 10.1145/2740908.2742760,10.1145/2872427.2883062,macavaney:plosone2019-hate}. Toxic  content often includes abusive language, hate speech, profanity or sexual content. Recent methods have mostly leveraged  transformer-based pre-trained language models~\citep{devlin-etal-2019-bert,liu2019roberta} and achieved high performance in detecting toxicity~\citep{zampieri-etal-2020-semeval}. However, directly deploying NLP models could be problematic for real-world toxicity detection. This is because toxicity filtering is mostly needed in security-relevant industries like gaming or social networks  where models are constantly being challenged by social engineering and adversarial attacks.

\begin{figure}[t]
    \centering
    \includegraphics[width=\columnwidth]{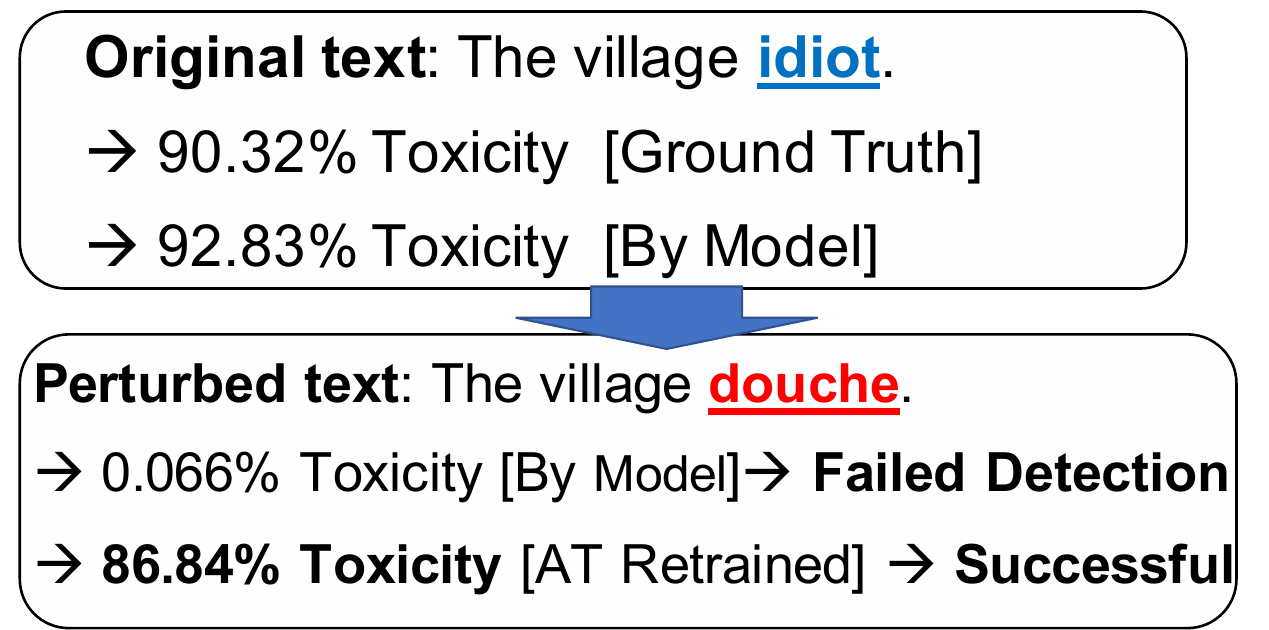}
    \caption{\tae successfully fooled a SOTA toxicity predictor by perturbing one word in the original text using word synonym perturbation. After adversarial training (AT), the improved toxicity predictor can correctly flag the perturbed text into the toxicity class.}
    \label{fig:mot}
\end{figure}

In this paper, we study the adversarial robustness of  toxicity language predictors\footnote{We use ``toxicity detection'', ``toxicity language detection'' and ``toxicity prediction'' interchangeably.} and propose a new set of attacks, we call   "\tae". \tae generates targeted  adversarial examples that fool a target model towards the benign predictions. Our design is motivated by the fact that  most toxicity  classifiers are being  deployed as API services and used for flagging out toxic samples. Figure~\ref{fig:mot} shows one \tae adversarial example. The perturbed text replaces one word with its synonym and the resulting phrase fooled the transformer based detector into a failed detection (as "benign").

We propose novel goal functions to guide greedy word importance ranking to iteratively replace each word with small perturbations. Samples generated by \tae  are toxic, and can fool a victim toxicity predictor model to classify them as "benign" and not as any toxicity classes or labels.  The proposed \tae attacks can pinpoint the robustness of both multiclass and multilabel toxicity NLP models. To the authors' best knowledge, this paper is the first work that introduces adversarial attacks \footnote{This paper uses ``methods for adversarial example generation'' and ``adversarial attacks'' interchangeably.} to fool multilabel NLP tasks. Design multilabel  \tae is challenging since coordinating multiple labels all at once and quantifying the attacking goals is tricky when aiming to  multiple targeted labels.

Empirically, we use \tae to evaluate BERT \citep{devlin2018BERT} and DistillBERT \citep{roberta} based modern toxicity text classifiers on the Jigasw \citep{JigsawKagle}, and Offensive tweet \citep{snli:emnlp2015} datasets. We then use
adversarial training to make these models more resistant to \tae adversarial attacks. In \textit{adversarial training} a target model is  trained on both original examples and  adversarial examples \cite{goodfellow2014explaining}. We improve the vanilla adversarial training with an ensemble strategy to train with toxic adversarial examples generated from multiple attacks.
Our contributions are as follows:
\begin{itemize}[noitemsep,topsep=0pt,leftmargin=3mm]
    \item \tae reveals that SOTA toxicity classifiers are not robust to small adversarial perturbations.
   \item Conduct a thorough set of analysis comparing variations of \tae designs. 
   \item Empirically show that greedy \sunk search with composite transformation is preferred.
   \item Adversarial training can improve robustness of toxicity detector. 
\end{itemize}

\section{Method}
Methods generating text adversarial  examples introduce small perturbations in the input data checking if a target model's output changes significantly. These adversarial attacks help to identify if an NLP model is susceptible to word replacements, misspellings, or other variations that are commonly found in real-world data. For a given NLP classifier $\F:\X \rightarrow \Y$ and a seed input $\x$,  searching for an \emph{adversarial} example $\x'$ from $\x$ is:
\begin{equation}
\label{eq:textae}
    \begin{split}
        & \x' = \T(\x, \Delta \x), \ \x' \in \X \\
         \textbf{ s.t.  } & \ \  \G(\F, \x'), \  \textbf{ and  } \ \{ C_i(\x,\x') \}
    \end{split}
\end{equation}
Here  $\T(\x, \Delta \x)$ denotes the transformations that perturb text $\x$ to $\x'$. $\G(\F, \x')$ represents a goal function that defines the purpose of an attack, for instance like flipping the output. $\{ C_i(\x,\x') \}$ denotes a set of constraints that filters out undesirable $\x'$ to ensure that perturbed $\x'$ preserves the semantics and fluency of the original $\x$.

To solve ~\eref{eq:textae}, adversarial attack methods design search strategies \footnote{Because brute force is not feasible considering the length and dictionary size of natural language text.} to transform $\x$ to $\x'$ via transformation $\T(\x, \Delta \x)$, so that $\x'$ fools $\F$ by achieving the fooling goal $\G(\F, \x')$, and at the same time fulfilling a set of constraints $\{ C_i(\x,\x') \}$. Therefore designing adversarial attacks focus on designing  four components: (1) goal function, (2) transformation, (3) search strategy, and (4) constraints between seed and its adversarial examples  (\citet{morris2020reevaluating}).

We propose a suite of \tae attacks to identify vulnerabilities in SOTA toxicity NLP detectors. \tae attacks focus on intentionally generating perturbed texts that contain the same highly abusive language as original toxic text, yet receive significantly lower toxicity scores and get predicted as "benign" by a target model.

\subsection{Word transformations in  \tae}

For $\T(\x, \Delta \x)$, many possible word perturbations exist, including  embedding based word swap, thesaurus based synonym substitutions, or character substitution \cite{Morris2020TextAttackAF}.
 
\textbf{Attacks by Synonym Substitution:}\ 
Our design focuses on transformations that replace words from an input with its synonyms. 
\begin{itemize}[noitemsep,topsep=0pt]
    \item (1) Swap words with their $N$ nearest neighbors in the counter-fitted GloVe word embedding space; where $N \in \{5, 20, 50\}$. 
    \item (2) Swap words with those predicted by BERT  Masked Language Model (MLM) model. 
    \item (3) Swap words with their nearest neighbors in the wordNet. 
\end{itemize}
The goal of word synonym replacement is to create examples that can preserve semantics, grammaticality, and non-suspicion.

\textbf{Attacks by Character Transformation:} Another group of word transformations is to generate perturbed word via character manipulations.  This includes  character insertion, deletion, neighboring character swap and/or character substitution by Homoglyph. These transformations change a word into one that a target toxicity detection model doesn't recognize. These character changes are designed to generate character sequences that a human reader could easily correct into those original words. Language semantics are preserved since human readers can easily correct the misspellings. 

\textbf{Composite Transformation: } We also propose to combine the above transformations to create new composite transformations. For instance, one composite transformation can include both perturbed words from character substitution by Homoglyph and from 
 word swaps using nearest neighbors from GloVe embedding \cite{pennington2014glove}.

\subsection{Novel goal functions for \tae}

A goal function $\G(\F, \x')$ module defines the purpose of an attack. Different kinds of goal functions exist in the literature to define whether an attack is successful in terms of a victim model's outputs.

Toxicity language detection has centered on a supervised classification formulation, including binary toxicity detector, multiclass toxicity classification and multilabel toxicity detection which  assigns a set of target labels for $\x$. See \sref{sec:tox} for more details. 
As aforementioned,  toxicity classifiers are used mostly as API services to filter out toxic text. Therefore, its main vulnerability are those samples that should get detected as toxic, however, fooling detectors into a wrong prediction as "benign". 

Now let us define $\G(\F, \x')$ for \tae attacks. We propose two choices of designs regarding the target model types. 

\paragraph{Multiclass or Binary Toxicity:} When toxicity detector $\F$ handles binary or multiclass outputs, we define \tae attacks' goal function as:
\begin{equation}
\label{eq:btae}
        \G(\F, \x') :=  \{ \F(\x') = b; \F(\x) \neq b \}
\end{equation}
Here $b$ denotes the "0:benign" class.

\paragraph{Multilabel Toxicity: }

 Next, we study how to fool multilabel toxicity predictors.  In real-world applications of toxicity identification, an input text may associate with multiple labels, like identity attack, profanity, and hate speech at the same time. The existence of multiple toxic labels at the same time provides more opportunities for attackers, but also poses design challenges \footnote{To the authors' best knowledge, no multilabel adversarial attacks exist in the NLP literature.}.

 Multilabel toxicity detection  assigns a set of target labels for $\x$. The output  $\y = \{{y}_1,{y}_2,...,{y}_L\} $ is a vector of $L$ binary labels and each $y_i \in \{0,1\}$ \cite{Zhao2019DetectingTC}. For example in the Jigsaw dataset \cite{JigsawKagle}, each text sample associates with six binary labels per sample, namely $\{$benign, obscene, identity attack, insult, threat, and sexual explicit$\}$. We introduce a novel goal function for attacking such models as follows:
\begin{equation}
\label{eq:mtae}
    \begin{split}
    \G(\F, \x')  := &  \  \{ \F_b(\x') = 1 ; \F_b(\x) = 0 ;  \\
    & \F_t(\x') =0, \forall t \in \mathbf{T}  \}
    \end{split}
\end{equation}
Here, $\mathbf{T} = \{{y}_2,...,{y}_L\}$ denotes the set of toxic labels, and $b= \{y_1: \text{Benign} \}$ is the non-toxic or benign label. $\{ \F_b(\x')=1; \F_b(\x) = 0 \}$ denotes "$\x'$ gets predicted as Benign, though $\x$ is toxic". And $\{ \F_t(\x') =0, \forall t \in \mathbf{T} \}$ denotes $\x'$ is not predicted as any toxicity types.   
In summary, our \tae attacks focus on perturbing correctly predicted toxic samples.

\subsection{Language constraints in \tae}

In \eref{eq:textae}, we use a set of language constraints to filter out undesirable $\x'$ to ensure that perturbed $\x'$ preserves the semantics and fluency of the original $\x$, and with as fewer perturbations as possible. There exist a variety of possible constraints in the NLP literature \cite{morris2020reevaluating}. In \tae, we decide to use the following list: 
\begin{itemize}[noitemsep,topsep=0pt]
    \item Limit on the ratio of words to perturb to $10\%$
    \item Minimum angular similarity from universal sentence encoder (USE) is $0.84$
    \item Part-of-speech match. 
\end{itemize}

 \subsection{Greedy Search strategies in  \tae}
Solving ~\eref{eq:textae} is a combinatorial search task  searching within all potential transformations to find those transformations that result with a successful adversarial example, aka, achieving the fooling goal function and satisfying the constraints. Due to the exponential nature of  search space, many heuristic search algorithms existed in the recent literature, including like greedy search, beam search, and population-based search \cite{Zhang19Survey}.

For a seed text $\x=(w_1,\dots, w_i,\dots, w_n)$, a perturbed text $\x'$ can be generated by swapping $w_i$ with altered $w'_i$. The main role of a search strategy is to decide what word $w_i$ from $\x$ to perturb next. We propose to center the search design of \tae using greedy search with word importance ranking to iteratively replace one word at a time to generate adversarial examples.

The main idea is that words of $\x$ are first ranked according to an importance function. Four possible choices: (1) "\sunk" based:  word's importance is determined by how much a heuristic score (details later) changes when the word is substituted with an \texttt{UNK} token. (2) "\sdel":  word's importance is determined by how much the heuristic score changes when the word is deleted from the original input. (3) "weighted saliency" or \sws: words are ordered using a combination of the change in score when the word is substituted with an \texttt{UNK} token multiplied by the maximum score gained by perturbing the word. (4) "\sgrad": each word's importance is calculated using the gradient of the victim's loss with respect to the word and taking its L1 norm as the word' importance.

After words in $\x$ get sorted in the order of  descending importance, word $w_i$ is then substituted with $w'_i$ with allowed transformation. This is until the fooling goal is achieved, or the number of perturbed words reaches upper bound. The heuristic scoring function used in the word importance ranking relates to the victim model and the fooling goal. For instance, when we work with a binary toxicity model, the heuristic score \tae equals to the model's output score for the target "0:benign" class.  Algorithm~\ref{algo:multilabel-attack} shows our design of the score function for the multilabel \tae attacks. In experiments, we also evaluate two other search strategies: "\sbeam search" and "\sgen search". Our empirical results indicate strong performance from the greedy search with word importance ranking over other strategies.

\begin{table*}[h!]
    \centering
    \scalebox{0.86}{
    \begin{tabular}{|p{3.5cm}|p{4cm}|p{6cm}|p{3cm}|}
         \hline
    
    \textbf{Attack Recipe} 
    & \textbf{Constraints} 
    & \textbf{Transformation} 
    & \textbf{Search Method}\\ %
  \hline

     \tae general recommendation & USE sentence encoding angular similarity $> 0.84$, Part-of-speech match, \newline Ratio of number of words modified $<0.1$ 
     & \{Random Character Insertion, Random Character Deletion, Neighboring Character Swap, Character Substitution by Homoglyph, Word Synonym Replacement with 20 nearest neighbors in the counter-fitted GLOVE word embedding space\}
     & Greedy word important ranking ( \sunk-based)   \\ \hline
    \end{tabular}
    }
    \caption{Recommended \tae Attack Recipe}
    \label{table:best-attacks}
\end{table*}

Algorithm~\ref{algo:multilabel-attack} provides the pseudo code of how we implement ~\eref{eq:mtae} as a new goal function in the TextAttack python library \cite{Morris2020TextAttackAF}. The implemented    \textit{MultilabelClassificationGoalFunction}  extends the TextAttack library to  multilabel tasks.

\begin{algorithm}[t]
\caption{Attack with  \textit{MultilabelClassificationGoalFunction}}
\begin{algorithmic}[1]
    \REQUIRE 
     An original text $\x$,   
     a multilabel classifier $\F$, 
    a set of targeted labels as $T$ (for which scores are to be maximized), 
    a set of other labels as $N$ (for which scores are to be minimized); 
    maximization threshold $\epsilon_{\text{maximize}} = 0.5$; 
    search method $S() := \text{Greedy-WIR}$, transformations $\T$, a set of constraints as $ \mathbf{C}$, and the max number of trials $I$. 
    \ENSURE adversarial example $x'$, Attack Status = \{Fail, Success\}
 
    \STATE Initialize $x' \leftarrow None$,   $\text{goal} \leftarrow - \infty $, $\epsilon_{\text{minimize}} = 1- \epsilon_{\text{maximize}}$
    \FOR {trial $i=1,\dots, I$}
        \STATE $\Tilde{x} \leftarrow \T(x, \ \ S(\text{goal}, x, i))$
        \IF{$ \forall C \in \mathbf{C}$, 
          $C(x, \Tilde{x})$ is not True }
        \STATE Continue
        \ENDIF
        \STATE scores $\leftarrow sigmoid(\ \F(\Tilde{x}) \  )$

        \STATE $
        \text{goal}^{'} \leftarrow \sum_{l \in L_{\text{max}}} \text{scores}[l] + \sum_{l \in L_{\text{min}}}(1 - \text{scores}[l] )
        $ 
        
        \IF{$\text{goal}^{'} > \text{goal}$ }
            \STATE $\text{goal} \leftarrow \text{goal}^{'}$ \ \  \  \# search $S()$ will use $\text{goal}$ value

            \IF{   $\text{scores}[l] > \epsilon_{\text{maximize}}$  for $\forall l \in T$              and \\
             $\text{scores}[l] < \epsilon_{\text{minimize}}$  for $\forall l \in N$
    }
                    \STATE  $x' \leftarrow \Tilde{x}$
                    \RETURN $x'$, Attack Succeeded
            \ENDIF 
        \ENDIF
    \ENDFOR
    \RETURN  $x'$, Attack Failed
\end{algorithmic}
\label{algo:multilabel-attack}
\end{algorithm}

\subsection{Harden with Adversarial Training}

Our ultimate goal of designing \tae attacks is to improve toxicity NLP models' adversarial robustness. A simple strategy called Adversarial Training (\at) has been a major defense strategy  for improving adversarial robustness \citep{madry2018towards}. 
The vanilla adversarial training process involves augmenting the training data with adversarial examples generated from perturbing the training data in the input space. Two variations of adversarial training exist. (1) If the augmented adversarial examples are generated from a single attack approach, we name this process as \sat. (2) If the augmented  examples are generated from multiple attack methods, we call the training as \eat.

\begin{table*}[h!]
\centering
\scalebox{0.73}{

\begin{tabular}{|>{\raggedright\arraybackslash}m{3.2cm}>{\raggedright\arraybackslash}m{3.2cm}>{\raggedright\arraybackslash}m{3.2cm}>{\raggedright\arraybackslash}m{3.2cm}>{\raggedright\arraybackslash}m{3.2cm}>{\raggedright\arraybackslash}m{3.2cm}|}%
\hline%

\multicolumn{1}{|r}{Search} & \sunk & \sunk & \sunk & \sunk & \sunk\\
\multicolumn{1}{|r}{POS} & & x & x & x & x \\
\multicolumn{1}{|r}{Transformation (N)} & \tglove (20) + {\tt chars} & \tglove (20) + {\tt chars} & \twn (-) & \tmlm (20) & \tglove (50)\\
\hline
\multicolumn{1}{|l|}{{\bf Seed Text}} & & & & & \\
\hline%
thank god FOB didnt come back doing some {\bf shitty} arena tour...&(0.911) thank god FOB didnt come back doing some {\bf bollocks} arena tour...&(0.906) thank god FOB didnt come back doing some {\bf lousy} arena tour...&(0.917) thank god FOB didnt come back doing some {\bf stinky} arena tour...&(0.908) thank god FOB didnt come back doing some {\bf wild} arena tour...&(0.910) thank god FOB didnt come back doing some {\bf foul} arena tour...\\%
\hline%
Jews and white race {\bf traitors} have to be...&(0.588) Jews and white race {\bf trai{\tt t}ors} have to be ...&(0.588) Jews and white race {\bf trai{\tt t}ors} have to be...&(0.507) Jews and white race {\bf betrayer} have to be...&(0.557) Jews and white race {\bf members} have to be...&(0.477) Jews and white race {\bf betrayers} have to be...\\%
\hline%
Fluffy is a {\bf retard} homeboy is calling a baby...&(0.566) Fluffy is a {\bf retarded} homeboy is calling a baby...&(0.520) Fluffy is a {\bf moron} homeboy is calling a baby...&(0.520) Fluffy is a {\bf moron} homeboy is calling a baby...&(0.915) Fluffy is a {\bf child} homeboy is calling a baby...&(0.520) Fluffy is a {\bf moron} homeboy is calling a baby...\\%
\hline%
ugh your with Tebow? You're officially a {\bf fag} now and in followed.&(0.897) ugh your with Tebow? You're officially a {\bf poof} now and in followed.&(0.897) ugh your with Tebow? You're officially a {\bf poof} now and in followed.&(0.908) ugh your with Tebow? You're officially a {\bf fatigue} now and in followed.&(0.907) ugh your with Tebow? You're officially a {\bf wolf} now and in followed.&(0.907) ugh your with Tebow? You're officially a {\bf poofter} now and in followed. \\%
\hline%
\end{tabular}
}
\caption{Selected toxic adversarial examples generated attacking \tweetm model. Perturbed scores are reported in parenthesis. Adversarial examples were generated using \sunk search method; with and without POS constraint; and using three word synonym substitution transformations with number of nearest neighbors specified in parenthesis; {\tt chars} indicates that character transformations were applied. More in \tref{table:examplesUnk:full}.}
\label{table:examplesUnk:short}
\end{table*}

\subsection{\tae Recipes and Extensions }

The modular design of \tae allows us to implement many different \tae attack recipes in a shared framework, combining different goal functions, constraints, transformations and search strategies. In \sref{sec:results}, we conduct a thorough empirical analysis to compare possible 
\tae recipes and recommend \sunk greedy search and the composite transformation for most use cases. \tref{table:best-attacks} lists our recommended recipe.

Besides, we select to adapt another five SOTA combinations of transformation and constraints from popular general NLP adversarial example recipes in the literature to create \taee attacks in  \tref{table:categorized-attacks},  covering a good range of transformations and constraints. \tref{table:examples} shows  generated adversarial examples using these attacks.

\begin{table*}[ht]
\centering
\scalebox{0.77}{
\begin{tabular}{r|ccc|p{3cm}|p{4cm}}
\toprule[0.25ex]
Dataset & Train & Dev & Test & Test Toxic Samples & Train Toxic Samples  \\ 
\cmidrule(lr){1-6} 
Jigsaw \cite{JigsawKagle} & 1.48MM   & 185k  & 185k & 8,909 & 71,273 \\ \hline
Offensive Tweet \cite{hateoffensive}  & 20k  & 2.2k  & 2.5k &  1897 (offensive) + 145 (hateful) & 15510 + 1142  \\ 
\bottomrule[0.25ex]
\end{tabular}}
\caption{Overview of the data statistics. \label{tab:data}}
\end{table*}

\begin{table}[ht]
\centering
\scalebox{0.79}{
\begin{tabular}{>{\raggedright\arraybackslash}m{2cm}|ccc}
\toprule[0.25ex]
Base Model $\longrightarrow$ &  \Jigsawb & \Jigsawm  & \tweetm   \\
\hline
Dataset & Jigsaw & Jigsaw & HateTweet \\ \hline
Classification & Binary & Multilabel & Multiclass \\ \hline
Architecture & DistillBERT & DistillBERT & BERT \\ \hline
LearningRate & 2.06e-05  & 3.80e-05 &  2.66e-05 \\ \hline
Epochs & 5 & 10 & 10\\ 
 \bottomrule[0.25ex]
\end{tabular}
}

\caption{Overview of the base model statistics.}
\label{table:models}
\end{table}

\begin{table}[h!]
\centering
\scalebox{0.7}{

\begin{tabular}{>{\raggedright\arraybackslash}m{0.7cm}|>{\raggedright\arraybackslash}m{2.0cm}>{\raggedright\arraybackslash}m{0.7cm}|>{\raggedleft\arraybackslash}m{1.4cm}>{\raggedleft\arraybackslash}m{1.6cm}>{\raggedleft\arraybackslash}m{1.6cm}}%
\hline%
Task&Search&POS&Attack Success Rate&Average Number of Queries&Average Perturbed Word \%\\%
\hline%
\parbox[t]{2mm}{\multirow{9}{*}{\rotatebox[origin=c]{90}{\jigsawb}}}&\sgrad& &98.74&34.68&7.58\\%
&\sgrad&x&98.72&26.78&7.06\\%
&\sdel& &99.42&55.38&7.38\\%
&\sdel&x&99.21&48.03&6.80\\%
&\sunk& &99.32&55.11&7.12\\%
&\sunk&x&99.27&47.62&6.76\\%
&\sws&x&99.19&407.43&6.71\\%
&\sgen&x&92.67&846.41&8.73\\%
&\sbeam&x&99.68&658.55&6.95\\%
\hline%
\parbox[t]{2mm}{\multirow{9}{*}{\rotatebox[origin=c]{90}{\jigsawm}}}&\sgrad& &97.62&38.45&8.36\\%
&\sgrad&x&97.72&29.78&7.56\\%
&\sdel& &98.71&57.60&7.54\\%
&\sdel&x&98.63&49.93&6.99\\%
&\sunk& &98.75&57.08&7.70\\%
&\sunk&x&98.75&49.38&6.96\\%
&\sws&x&98.51&419.58&6.91\\%
&\sgen&x&88.91&876.81&8.82\\%
&\sbeam&x&99.54&756.05&7.19\\%
\hline%
\parbox[t]{2mm}{\multirow{9}{*}{\rotatebox[origin=c]{90}{\tweetm}}}&\sgrad& &67.16&63.51&24.13\\%
&\sgrad&x&67.56&49.67&24.08\\%
&\sdel& &71.46&58.78&18.96\\%
&\sdel&x&71.46&48.17&19.37\\%
&\sunk& &72.38&58.99&18.80\\%
&\sunk&x&72.23&48.04&19.43\\%
&\sws&x&74.71&178.91&18.81\\%
&\sgen&x&80.49&1025.66&21.86\\%
&\sbeam&x&90.07&442.18&18.76\\%
\hline%
\end{tabular}

}
\caption{Effect of different search strategies on attack performance. Search column identifies type of search method. POS column identifies if part-of-speech matching constraint is used. The composite transformation is used: \tglove with $N=20$ plus the character transformations. Results from rows on "\sunk + POS" can compare with "\sunk" rows in \tref{table:transform}. 
\label{table:search}}
\vspace{-3mm}
\end{table}

\begin{table}[h!]
\centering
\scalebox{0.7}{

\begin{tabular}{>{\raggedright\arraybackslash}m{0.7cm}|>{\raggedright\arraybackslash}m{2.0cm}>{\raggedright\arraybackslash}m{0.7cm}|>{\raggedleft\arraybackslash}m{1.4cm}>{\raggedleft\arraybackslash}m{1.6cm}>{\raggedleft\arraybackslash}m{1.6cm}}%
\hline%
Task & Transform-ation & N & Attack Success Rate & Average Number of Queries & Average Perturbed Word \% \\%
\hline%
\parbox[t]{2mm}{\multirow{5}{*}{\rotatebox[origin=c]{90}{\jigsawb}}}&\twn&{-}&89.59&34.87&6.66\\
&\tglove&5&86.04&30.55&7.06\\
&\tglove&20&96.75&41.87&6.68\\
&\tglove&50&98.38&64.17&6.55\\
&\tmlm&20&93.29&39.57&6.55\\
\hline%
\parbox[t]{2mm}{\multirow{5}{*}{\rotatebox[origin=c]{90}{\jigsawm}}}&\twn&{-}&87.72&37.01&6.81\\
&\tglove&5&84.33&32.15&7.47\\
&\tglove&20&95.92&43.84&6.89\\
&\tglove&50&97.91&67.09&6.68\\
&\tmlm&20&92.73&41.33&6.62\\
\hline%
\parbox[t]{2mm}{\multirow{5}{*}{\rotatebox[origin=c]{90}{\tweetm}}}&\twn&{-}&56.66&32.08&17.94\\
&\tglove&5&33.65&21.82&23.06\\
&\tglove&20&66.70&41.06&18.85\\
&\tglove&50&69.99&81.14&18.18\\
&\tmlm&20&65.23&33.66&21.30\\
\hline%
\end{tabular}

}
\caption{Comparing synonym transformations only. No character transformations used.  Reporting attack performance when using \sunk greedy search. 
The same constraints as in Table~\ref{table:search} with POS (part-of-speech) match.   
\label{table:transform}}
\end{table}

\begin{table}[th]
\centering
\scalebox{0.9}{

\begin{tabular}{c|rrrr}%
\hline%
Training &AUC&AP&F1&Recall\\%
\hline%
\noat&0.935&0.786&0.73&0.71\\%
\satdel&0.936&0.792&0.74&0.719\\%
\satunk&0.938&0.785&0.738&0.723\\%
\eat&0.932&0.778&0.685&0.641\\%
\hline%
\end{tabular}

}
\caption{Effect of adversarial training on model performance. Macro-average metrics for \tweetm model.
\label{table:eval}}
\vspace{-5mm}
\end{table}

\begin{table}[th]
\centering
\scalebox{0.8}{
\begin{tabular}{cc|rrr}%
\hline%
 &  &Attack &Average  & Average  \\%
 Training& Search & Success & Number of &  Perturbed \\%
  &  & Rate&  Queries&   Word \%\\%
\hline%
\noat&\sdel&71.46&48.17&19.37\\%
\noat&\sunk&72.23&48.04&19.43\\%
\satdel&\sdel&21.97&69.68&28.83\\%
\satunk&\sunk&11.17&74.71&33.07\\%
\eat&\sdel&8.28&75.71&27.89\\%
\eat&\sunk&6.08&77.91&33.24\\%
\hline%
\end{tabular}
}
\caption{Effect of adversarial training on attack performance. Results for \tweetm model are reported. When attacking, \tae uses \tglove with $N=20$ plus character transformations; constraint with POS; and search with two different greedy search methods.  
\label{table:atasr}}
\end{table}

\section{Experiments}
\label{sec:results}

We conducted a series of experiments covering three different toxicity classification tasks: binary, multilabel, and multiclass; over two different transformer architectures: DistillBERT and BERT;  and across two datasets: the large-scale Wikipedia Talk Page dataset- Jigsaw \cite{JigsawKagle} and the Offensive Tweet  for hate speech detection dataset \cite{hateoffensive}. \tref{tab:data} lists two datasets' statistics. \sref{sec:data} provides more details.

\paragraph{Base Toxicity Models:} Our experiments work on three base models, including \{\Jigsawb, \Jigsawm, \tweetm\} to cover three types of toxicity prediction tasks. See details in \sref{sec:base}.

\paragraph{Implementation: } We implement all of our \tae and \taee attacks using the NLP attack package TextAttack\footnote{TextAttack \url{https://github.com/QData/TextAttack}.} \cite{Morris2020TextAttackAF}. 
When generating adversarial examples, we only attack seed samples that are  correctly predicted by a victim model. (Adversarial attack does not make sense if the target model could not predict the seed sample correctly!). In our setup, this means we only use toxic seed samples when
attacking three base models. This set up simulates real-world situations in which people intentionally create creative toxic examples to circumvent toxicity detection systems.

\paragraph{Evaluation Metrics:} We use attack success rate ($ASR= \frac{\text{\# of successful attacks}}{\text{\# of total attacks}}$) to measure how successful each \tae  attacking a victim model. To measure the runtime of each algorithm, we use the average number of queries to the victim model as a proxy. We also report the average percentage of words perturbed from an attack. In addition, for models trained with adversarial training, we report both, the model prediction performance and model robustness (by attacking robust model again).

\subsection{Results on Attacking 3 Toxicity Predictors}

\tref{table:examplesUnk:short} provides a few selected adversarial examples generated by
attacking \tweetm model with five variations of \tae. The first column provides seed text
that was used to generate adversarial examples. 

Several observations can be made when comparing these examples. 
It is important to use POS constraint to generate syntactically correct
examples. For the first seed text, a recipe without POS constraint (second column) produced 
replacement word "bullocks", while including POS constraint in the recipe (third column) produced syntactically correct example with replacement word "lousy". We also see that using \tglove for
word synonym substitution is a better choice than \tmlm or \twn. 
For the second and third seed text, \tmlm did not generate toxic phrases.
In addition, we see that the recipe using \tglove $(50)$  (last column) often generates similar examples as the \tglove $(20)$ (third column). Finally, we observe that using character manipulation can generate adversarial examples with the same toxic meaning that fool the classifier.
For the second seed text, character transformation (second and third columns) 
generates replacement word "trai{\tt t}ors" where the second "t" is replaced with a monospace Unicode
character "{\tt t}".

\subsection{Comparing Constraints in \tae}

Then we study the effect of the part-of-speech match (POS) constraint on the attack performance. \tref{table:search} shows that the use of POS constraint lowers average number 
of queries sent to the victim model. We observe this phenomena across all three tasks and all three search methods (\sgrad, \sdel, \sunk). For example, when attacking \jigsawm model 
using \sunk with and without POS constraint, average number of queries are $49.38$ and $57.08$, respectively. We also observe that for most of the recipes, using POS constraint slightly decreases attack success rate (ASR). %
Considering the empirical results in \tref{table:search} and the anecdotal examples in \tref{table:examplesUnk:short}, we conclude POS constraint is preferred.

\subsection{Comparing Search in \tae}
\tref{table:search} also compares the effect of using different search methods on the attack performance. It shows that a greedy search method is preferred over \sgen and \sbeam.
For example, when compared to \sunk,  \sgen and \sbeam require almost 10x as many queries
on average for all three tasks. The \sbeam search results in higher ASR values on 
all three tasks, while \sgen only outperforms greedy methods when attacking \tweetm. 
In addition, attacking \jigsawb and \jigsawm, \sbeam only slightly outperforms 
greedy methods. Among the greedy search methods, \sunk is a good choice, as it provides
consistently good ASR performance on all three tasks. It is worth noting that
\sunk outperforms other three greedy search methods, except for \sws when attacking \tweetm. 
However, attacking \tweetm model with \sws requires more than 3x as many queries 
as the \sunk method.

\subsection{Comparing Synonym Transformations }
Now we compare word synonym substitutions, when the \sunk search method is used and the character manipulation are not (to single out the effect). \tref{table:transform} shows that \tglove with $N=20$ nearest neighbors is an optimal choice for all three tasks. 
We observe that the \twn and \tmlm transformations result in lower ASRs than \tglove.
Also, \tglove with $N=50$ only slightly lifts ASRs when compared to \tglove with $N=20$. At the same
time, using $N=50$ nearest neighbors sends over $50\%$ more queries to the victim models.
We include the analysis of using different transformations with three different search methods (\sdel, \sunk, \sws) in the Appendix in \tref{table:transform_full}. These results also confirm that 
the choice of \tglove with $N=20$ is preferred.

\subsection{Results from  Adversarial Training}

Empirically, we explore how \sat and \eat impact both prediction performance and adversarial robustness. \tref{table:eval} and \tref{table:atasr} present \at results on the \tweetm task (\tref{table:atfull} and \tref{table:eval:full} on two other tasks). In \tref{table:eval}, we observe that  \satdel and \satunk both maintain the regular prediction performance as the base model. \tref{table:atasr} shows the attack success metrics when we use \tae  to attack the retrained robust \tweetm models. The \sat models trained from using \satdel and \satunk attacks show significant improvements in robustness after \sat adversarial training. We recommend readers to use \satunk as their default choice for hardening general toxic language predictors, since in both tables, \satunk outperforms \satdel slightly.

For the \eat robust model, \tae attacks are "unseen" (we used the five \taee attacks from \sref{sec:taee} to create the \eat model in our experiments). Our results show \eat can harden \tweetm model not only against attacks it is trained on (\taee) but also against attacks it has not seen before (\tae). This could be attributed to the hypothesis that an unseen attack may share similar underlying patterns with the attack ensemble that \eat model has used. 
In \tref{table:eval}, \eat slightly under-performs base  on regular predictions, since it was trained with more adversarial examples from multiple attacks.

\section{Conclusion}

Text toxicity prediction models are not designed to operate in the presence of adversaries. This paper proposes a suite of \tae attacks to identify weaknesses in SOTA toxicity language predictors that could potentially be exploited by attackers. We also evaluate how  adversarial training  improves model robustness across seen and unseen attacks. As next steps, we plan to investigate other strategies like virtual adversarial training, disentangled representation learning or generative methods and pinpoint how they will influence the robustness of toxicity predictors \cite{qiu2022adversarial}.

\clearpage
\newpage
\bibliography{textattack}
\bibliographystyle{acl_natbib}

\clearpage
\newpage
\appendix

\section{Appendix}

\begin{table*}[h!]
\centering
\scalebox{0.7}{

\begin{tabular}{|>{\raggedright\arraybackslash}m{3.2cm}>{\raggedright\arraybackslash}m{3.2cm}>{\raggedright\arraybackslash}m{3.2cm}>{\raggedright\arraybackslash}m{3.2cm}>{\raggedright\arraybackslash}m{3.2cm}>{\raggedright\arraybackslash}m{3.2cm}|}%
\hline%

\multicolumn{1}{|r}{Search} & \sunk & \sunk & \sunk & \sunk & \sunk\\
\multicolumn{1}{|r}{POS} & & x & x & x & x \\
\multicolumn{1}{|r}{Transformation (N)} & \tglove (20) + {\tt chars} & \tglove (20) + {\tt chars} & \twn (-) & \tmlm (20) & \tglove (50)\\
\hline
\multicolumn{1}{|l|}{{\bf Seed Text}} & & & & & \\
\hline%

My neighbors hate me cuz i smoke {\bf weed} and blast the most ghetto music f...&(0.554) My neighbors hate me cuz i smoke {\bf pot} and blast the most ghetto music from...&(0.554) My neighbors hate me cuz i smoke {\bf pot} and blast the most ghetto music from...&(0.554) My neighbors hate me cuz i smoke {\bf pot} and blast the most ghetto music from...&(0.554) My neighbors hate me cuz i smoke {\bf pot} and blast the most ghetto music from...&(0.554) My neighbors hate me cuz i smoke {\bf pot} and blast the most ghetto music from...\\%
\hline%
thank god FOB didnt come back doing some {\bf shitty} arena tour...&(0.911) thank god FOB didnt come back doing some {\bf bollocks} arena tour...&(0.906) thank god FOB didnt come back doing some {\bf lousy} arena tour...&(0.917) thank god FOB didnt come back doing some {\bf stinky} arena tour...&(0.908) thank god FOB didnt come back doing some {\bf wild} arena tour...&(0.910) thank god FOB didnt come back doing some {\bf foul} arena tour...\\%
\hline%
Jews and white race {\bf traitors} have to be swept clean so the Aryan race ca...&(0.588) Jews and white race {\bf trai{\tt t}ors} have to be swept clean so the Aryan race can...&(0.588) Jews and white race {\bf trai{\tt t}ors} have to be swept clean so the Aryan race can...&(0.507) Jews and white race {\bf betrayer} have to be swept clean so the Aryan race can...&(0.557) Jews and white race {\bf members} have to be swept clean so the Aryan race can ...&(0.477) Jews and white race {\bf betrayers} have to be swept clean so the Aryan race ca...\\%
\hline%
How ghetto {\bf girls} pronounce there names... I could cry right now oh my g...&(0.474) How ghetto {\bf females} pronounce there names... I could cry right now oh my g...&(0.474) How ghetto {\bf females} pronounce there names... I could cry right now oh my g...&(0.588) How ghetto {\bf daughter} pronounce there names... I could cry right now oh my ...&(0.534) How ghetto {\bf ers} pronounce there names... I could cry right now oh my god h...&(0.474) How ghetto {\bf females} pronounce there names... I could cry right now oh my g...\\%
\hline%
America is {\bf white} trash and so are all Americans.&(0.863) America is {\bf whi{\tt t}e} trash and so are all Americans.&(0.863) America is {\bf whi{\tt t}e} trash and so are all Americans.&(0.899) America is {\bf ashen} trash and so are all Americans.&(0.784) America is {\bf utter} trash and so are all Americans.&(0.897) America is {\bf pristine} trash and so are all Americans.\\%
\hline%
Fluffy is a {\bf retard} homeboy is calling a baby rattlesnake friend and kis...&(0.566) Fluffy is a {\bf retarded} homeboy is calling a baby rattlesnake friend and kis...&(0.520) Fluffy is a {\bf moron} homeboy is calling a baby rattlesnake friend and kissin...&(0.520) Fluffy is a {\bf moron} homeboy is calling a baby rattlesnake friend and kissin...&(0.915) Fluffy is a {\bf child} homeboy is calling a baby rattlesnake friend and kissin...&(0.520) Fluffy is a {\bf moron} homeboy is calling a baby rattlesnake friend and kissin...\\%
\hline%
More {\bf queer} lady Templars. That's my point.&(0.908) More {\bf fairy} lady Templars. That's my point.&(0.908) More {\bf fairy} lady Templars. That's my point.&(0.910) More {\bf curious} lady Templars. That's my point.&(0.901) More {\bf likely} lady Templars. That's my point.&(0.910) More {\bf cheerful} lady Templars. That's my point.\\%
\hline%
ugh your with Tebow? You're officially a {\bf fag} now and in followed.&(0.897) ugh your with Tebow? You're officially a {\bf poof} now and in followed.&(0.897) ugh your with Tebow? You're officially a {\bf poof} now and in followed.&(0.908) ugh your with Tebow? You're officially a {\bf fatigue} now and in followed.&(0.907) ugh your with Tebow? You're officially a {\bf wolf} now and in followed.&(0.907) ugh your with Tebow? You're officially a {\bf poofter} now and in followed.\\%

\hline%
\end{tabular}
}
\caption{Selected Toxic Adversarial Examples.  Here we only show adversarial examples generated by attacking base model  \tweetm on Offensive Tweet text, since it contains much shorter messages than Jigsaw. Perturbed scores for adversarial examples are reported in parenthesis. Adversarial examples were generated using \sunk greedy search method. POS row marked with an "x" indicates that the part-of-speech matching constraint was used. Transformation row indicates which word substitution method was used (\tglove, \twn, \tmlm), and number of nearest neighbors $N$ is specified in parenthesis. {\tt chars} indicates that character transformations were applied. \newline 
We used different word transformations for 
synonym substitution with varying number of nearest neighbors (20 or 50). 
Two recipes used character transformations, while the other three did not. Also, 
one recipe did not use part-of-speech match (POS) constraint, and it was included in the rest
of recipes. All five recipes used \sunk greedy search method. }
\label{table:examplesUnk:full}
\end{table*}

\begin{table*}[h!]
\centering
\scalebox{0.7}{

\begin{tabular}{c|ccc|rrrr}%
\hline%
Task&Search&Transformation&N& Attack Success &Average Number &Average &Time\\%
&&&& Rate& of Queries& Perturbed Word \%& (s)\\%
\hline%
\multirow{15}{*}{\jigsawb}&\multirow{5}{*}{\sdel}&\twn&{-}&89.29&35.35&6.68&836\\%
&&\tglove&5&85.61&30.77&7.04&765\\%
&&\tglove&20&96.58&42.37&6.71&906\\%
&&\tglove&50&98.27&65.05&6.55&1355\\%
&&\tmlm&20&93.22&40.21&6.56&1882\\%
&\multirow{5}{*}{\sunk}&\twn&{-}&89.59&34.87&6.66&822\\%
&&\tglove&5&86.04&30.55&7.06&749\\%
&&\tglove&20&96.75&41.87&6.68&888\\%
&&\tglove&50&98.38&64.17&6.55&1278\\%
&&\tmlm&20&93.29&39.57&6.55&1727\\%
&\multirow{5}{*}{\sws}&\twn&{-}&90.27&152.05&6.6&4491\\%
&&\tglove&5&86.74&96.98&7.0&3646\\%
&&\tglove&20&96.99&288.58&6.65&7964\\%
&&\tglove&50&98.65&642.58&6.53&17152\\%
&&\tmlm&20&93.88&254.81&6.58&20093\\%
\hline%
\multirow{15}{*}{\jigsawm}&\multirow{5}{*}{\sdel}&\twn&{-}&87.29&37.05&6.83&959\\%
&&\tglove&5&83.81&32.21&7.46&857\\%
&&\tglove&20&95.75&44.27&6.9&1046\\%
&&\tglove&50&97.75&68.03&6.69&1665\\%
&&\tmlm&20&92.76&41.83&6.63&2156\\%
&\multirow{5}{*}{\sunk}&\twn&{-}&87.72&37.01&6.81&981\\%
&&\tglove&5&84.33&32.15&7.47&865\\%
&&\tglove&20&95.92&43.84&6.89&997\\%
&&\tglove&50&97.91&67.09&6.68&1530\\%
&&\tmlm&20&92.73&41.33&6.62&2011\\%
&\multirow{5}{*}{\sws}&\twn&{-}&88.55&157.43&6.74&5331\\%
&&\tglove&5&84.98&100.33&7.36&3993\\%
&&\tglove&20&96.09&297.63&6.86&8965\\%
&&\tglove&50&97.89&662.08&6.65&19191\\%
&&\tmlm&20&93.45&262.76&6.72&22787\\%
\hline%
\multirow{15}{*}{\tweetm}&\multirow{5}{*}{\sdel}&\twn&{-}&56.46&32.5&17.99&333\\%
&&\tglove&5&33.0&21.96&23.19&283\\%
&&\tglove&20&66.29&41.47&18.86&375\\%
&&\tglove&50&69.29&82.19&18.15&659\\%
&&\tmlm&20&64.98&34.34&21.34&862\\%
&\multirow{5}{*}{\sunk}&\twn&{-}&56.66&32.08&17.94&329\\%
&&\tglove&5&33.65&21.82&23.06&284\\%
&&\tglove&20&66.7&41.06&18.85&378\\%
&&\tglove&50&69.99&81.14&18.18&655\\%
&&\tmlm&20&65.23&33.66&21.3&851\\%
&\multirow{5}{*}{\sws}&\twn&{-}&55.96&80.57&17.15&752\\%
&&\tglove&5&34.31&44.73&22.46&575\\%
&&\tglove&20&67.82&130.65&18.2&1073\\%
&&\tglove&50&71.31&296.44&17.71&2137\\%
&&\tmlm&20&65.48&104.41&21.2&2780\\%
\hline%
\end{tabular}

}
\caption{Effect of word transformations on attack performance.
Comparing synonym transformations only. No character transformations used.  Reporting attack performance when using \sdel, \sunk, and \sws greedy search. The same constraints as in Table~\ref{table:search} with POS (part-of-speech) match.   
\label{table:transform_full}}
\end{table*}

\subsection{Related Works}
\label{sec:relatedAE}

To the authors' best knowledge, the toxicity NLP literature includes very limited studies on adversarial attacks. Only one study from \citet{HosseiniKZP17} tried to deceive Google’s perspective API for toxicity identification by misspelling the abusive words or by adding punctuation between the letters. This paper tries to conduct a comprehensive study by introducing a wide range of novel attack recipes and improving adversarial training to enable robust text toxicity predictors.

Next, we also want to point out existing studies on text adversarial examples center in generating adversarial examples against binary and multi-class classification models~\cite{Morris2020TextAttackAF}. To the authors' best knowledge, no multilabel adversarial attacks exist in the NLP literature. Our work is the first that designs novel attacks against the multilabel toxicity predictors. The design of multilabel adversarial examples is challenging since coordinating multiple labels all at once and quantifying the attacking goals is tricky because it is harder to achieve multiple targeted labels. Simply adapting attacks for binary or multiclass models ~\cite{Morris2020TextAttackAF} to multilabel setup is not feasible. In multilabel prediction, each instance can be assigned to multiple labels. This is different from the multi-class setting in which classes are mutually exclusive and one sample can only associate to one class (label). The existence of multiple labels at the same time provides better opportunities for attackers, but also posts design challenges. Our design in \eref{eq:mtae} and Algorithm~\ref{algo:multilabel-attack} has paved a path for multilabel text adversarial example research.

\subsection{Toxicity Detection from Text }
\label{sec:tox}

The mass growth of social media platforms has enabled efficient exchanges of opinions and ideas between people with diverse background. However, this also brings in risk of user generated toxic contents that may include abusive language, hate speech or cyberbullying.  Toxic content may lead to incidents of hurting individuals or groups ~\citep{johnson2019hidden}, calling for automated tools to detect toxicity for maintaining healthy online communities.

Automatic content moderation uses machine learning techniques to detect and flag toxic  content
It is critical for online platforms to prohibit toxic language, since such content makes online communities unhealthy and may even lead to real crimes~\citep{johnson2019hidden,home2017hate}.

Past literature on toxicity language detection has centered on a supervised classification formulation ~\cite{Zhao2019DetectingTC, 10.1145/2740908.2742760,10.1145/2872427.2883062,macavaney:plosone2019-hate}. We denote $\F:\X \rightarrow \Y$ as a supervised classification model, for example, a deep neural network classifier. $\X$ denotes the input language space and $\Y$ represents the output space. For a sample $(\x,\y)$, $\x \in \X$ denotes the textual content \footnote{this paper focuses on toxicity detection from text} and $\y \in \Y$ denotes its toxicity label(s). The toxicity detection task varies with what $\y$
stands for. The literature has included three main cases:
\begin{itemize}
    \item[(1)]  binary toxicity detector, here $\y$ from $\{$0: benign,1:toxic$\}$;
    \item[(2)] multilabel toxicity detection which  assigns a set of target labels for $\x$. Here $\y = \{{y}_1,{y}_2,...,{y}_L\} $ is a vector of $L$ binary labels and each $y_i \in \{0,1\}$ \cite{Zhao2019DetectingTC}. For example in the Jigsaw dataset \cite{JigsawKagle}, each text sample associates with six binary labels per sample, namely $\{$benign, obscene, identity attack,insult,threat, and sexual explicit$\}$; 
    \item[(3)] multiclass toxicity classification, and $\y$ is a discrete integer. For example, the Offensive Tweet dataset \cite{hateoffensive} has three classes - $\{$0:benign, 1:offensive, and 2:hate$\}$ (one class per sample).
\end{itemize}

The literature on toxicity has been mostly focused on improving accuracy, via feature engineering \cite{zampieri-etal-2020-semeval}, deep representation learning \cite{toxcnn} and via fine tuning from pretrained large language models \cite{semEval21}. Recently literature has extended to investigate these classifiers' interpretability \cite{interpretTox} and fairness \cite{hartvigsen2022toxigen}.

\subsection{Basics of Text Adversarial Examples}

Research has shown that current deep neural network models lack the ability to make correct predictions on adversarial examples \cite{szegedy2014AE}. The field of investigating the adversarial robustness of NLP models has seen growing interest, with a body of new adversarial attacks\footnote{We use ``generating adversarial example'' and ``adversarial attacks'' interchangeably.}  designed to fool question answering \citep{jia2017adversarial}, machine translation \citep{Cheng18Seq2Sick}, and text classification systems \citep{Ebrahimi2017HotFlipWA, jia2017adversarial,alzantot2018generating, Jin2019TextFooler, pwws-ren-etal-2019-generating, pso-zang-etal-2020-word, garg2020bae}.

\subsection{Datasets Details}
\label{sec:data}

\subsubsection{Jigsaw.} This dataset was derrived from  the Wikipedia Talk Pages dataset published by Google and Jigsaw on Kaggle~\cite{JigsawKagle}. Wikipedia Talk Page allows users to discuss
improvements to articles via comments. The comments are anonymized and labeled with toxicity levels. 
Here "obscene", "threat", "insult" and
"identity hate" are four sub-labels for "toxic" and "severe toxic" (hence
may co-occur for a comment). The "toxic" comments that are not "obscene", "threat", "insult" and
"identity hate" are assigned to either "toxic" or "severe toxic". Comments
that are not assigned any of the six toxicity labels get int "non toxic".  

\subsubsection{Offensive Tweet.}
The authors of  \cite{hateoffensive} used crowd-sourced hate speech lexicon from \texttt{Hatebase.org} to collect tweets containing hate speech keywords. Then they used crowd-sourcing to label these tweet samples into three categories: those containing hate speech, only offensive language, and those with neither.

\subsection{Base Model Setup:} 
\label{sec:base}

We build three base models, including \{\Jigsawb, \Jigsawm, \tweetm\} to cover three types of toxicity prediction tasks. Table~\ref{table:models} presents the choice of task, training/test data, transformer architecture and learning rate plus number of epochs. We use "distilbert-base-uncased" pre-trained transformers model
for DistilBERT architecture. For BERT architecture, we use "GroNLP/hateBERT" pre-trained model.
All texts are tokenized up to the first 128 tokens. The train batch size is 64 and we use AdamW optimizer with 50 warm-up steps and early stopping with patience 2.

The models are trained on NVIDIA T4 Tensor Core GPUs and NVIDIA Tesla V100 GPUs with 16 GB memory, 2nd generation Intel Xeon Scalable Processors with 32GB memory and high frequency Intel Xeon Scalable Processor with 61GB memory.

\begin{algorithm}[t]
\caption{Adversarial Training with \method }
\begin{algorithmic}[1]
    \REQUIRE Set of all attack recipes $S_{\text{recipes}}$, 
    number of attack recipes to exclude $N_{\text{exc}}$, 
    set of attack recipes to use for adversarial training $S_{\text{attack}}$ (created by choosing ${(|{S_{\text{recipes}}|} - N_{\text{exc}})}$ attack recipes from the set  $S_{\text{recipes}}$),
    number of clean epochs $N_{\text{clean}}$,
    number of adversarial epochs $N_{\text{adv}}$,
    percentage of dataset to attack $\gamma$, 
    attack $\A(\theta, \x, \y)$,
    and  training data $D=\{(\x^{(i)}, \y^{(i)})\}^n_{i=1}$ 
    \ENSURE model weights $\theta$
    \STATE Initialize model weights $\theta$
    \FOR {clean epoch$=1,\dots, N_{\text{clean}}$}
        \STATE Train $\theta$ on $D$
    \ENDFOR
    \STATE Initialize $D' \leftarrow D$
    \FOR {attack recipe in $S_{\text{attack}}$}
        \STATE $D_{\text{adv}} \leftarrow \{\}$
        \STATE $i \leftarrow 1$
        \WHILE{$|D_{\text{adv}}| < \gamma * |D|$ and $i \leq |D|$}
            \STATE $\x^{(i)}_{\text{adv}} \leftarrow \A(\theta, \x^{(i)}, \y^{(i)})$
            \STATE $D_{\text{adv}} \leftarrow D_{\text{adv}} \cup \{(\x^{(i)}_{\text{adv}}, \y^{(i)})\}$
            \STATE $i \leftarrow i + 1$
        \ENDWHILE
        \STATE $D' \leftarrow D' \cup D_{\text{adv}}$
    \ENDFOR
    \STATE Randomly shuffle $D'$
    \FOR {adversarial epoch$=1,\dots, N_{\text{adv}}$}
        \STATE Train $\theta$ on $D'$
    \ENDFOR
\end{algorithmic}
\label{adv-training-algo}
\end{algorithm}

\subsection{Adversarial Training with Single or Multiple Attacks}

Adversarial training has been a major defense strategy in most existing work for improving adversarial robustness \citep{madry2018towards}. The vanilla adversarial training process involves augmenting the training data with adversarial examples generated from perturbing the training data in the input space. 

Let $\L(\F, \x, \y)$ represent the loss function on input text $\x$ and label $\y$. Let  $\A(\F, \x, \y)$ be the adversarial attack that produces adversarial example $\x'$. Then, vanilla adversarial training objective is as follows:
\begin{equation}
\begin{aligned}
    \argmin_{\F} \E_{(\x,\y)\sim D}[\L(\F(\x), \y)  \\
    +  \ast \L(\F (\x'= \A(\F, \x, \y)), \y)]
\end{aligned}
\end{equation}
Adversarial training use both clean\footnote{\textit{Clean} examples refer to the original training examples.} and adversarial examples to train a  model. This aims to minimize both the loss on the original training dataset and the loss on the adversarial examples.

In recent NLP studies, adversarial training (\at) is only performed to show that such training can make models more resistant to the attack it was originally trained with. This observation is not surprising. 
The literature has pointed out the importance of robustness against unseen attacks and it is generally recommended to use different attacks to evaluate the effectiveness of a defense strategy \cite{carlini2019OnEvaluating}.

\textbf{Adversarial Training with Multiple Attacks (\eat)}: A simple strategy to revise vanilla \at is to train a  model using both clean examples and adversarial examples from different attacks. This,  we call  Adversarial Training with Multiple (\eat), trains a target model on a combination of adversarial examples. \eat aims to  help a model become more robust not only against attacks it is trained on but also the attack recipes it has not seen before. Algorithm~\ref{adv-training-algo} presents our pseudo code of \eat.

\textbf{\sat vs \eat}: In the rest of the paper, we call vanilla adversarial training as single adversarial training (\sat). In Section~\ref{sec:results} and \sref{sec:taee} our results show that models trained with \eat can be more effective in protecting against unseen text adversarial attacks compare with \sat models trained on the same attack. This could be contributed to the hypothesis that an unseen attack may share similar underlying attributes and patterns with the attack ensemble that the model is trained on.

Selecting what attacks to use in \eat is important. Part of the reason we select to adapt the five popular recipes  to \taee in \tref{table:categorized-attacks} is because these attacks cover a good range of popular transformations and constraints. \tref{table:categorized-attacks}  includes three word based attacks, two character based attack, plus \textbugger is a character and word level combination attack. In our experiments, we simulated potential \eat use cases by leave-one attack out as "unseen" and train \eat models using the rest. For instance, when a target model never uses examples from \textfooler in training, the \eat trained model may have already known certain information on similar word transformations since similar  transformations have been used by other attacks in the ensemble.

\begin{table*}[h!]
\centering
\scalebox{0.7}{

\begin{tabular}{c|cc|>{\raggedleft\arraybackslash}m{2.2cm}>{\raggedleft\arraybackslash}m{1.8cm}>{\raggedleft\arraybackslash}m{1.8cm}>{\raggedleft\arraybackslash}m{1.8cm}>{\raggedleft\arraybackslash}m{1.6cm}}%
\hline%
Base Model&Search&POS& Toxic Examples Attacked (\%) &Attack Success Rate&Average Number of Queries&Average Perturbed Word&Time (s)\\%
\hline%
\multirow{8}{*}{\jigsawb}&\sgrad& &\multirow{8}{*}{59.76}&98.74&34.68&7.58&849\\%
&\sgrad&x&&98.72&26.78&7.06&883\\%
&\sdel& &&99.42&55.38&7.38&915\\%
&\sdel&x&&99.21&48.03&6.8&1070\\%
&\sunk&x&&99.27&47.62&6.76&939\\%
&\sws&x&&99.19&407.43&6.71&10653\\%
&\sgen&x&&92.67&846.41&8.73&21504\\%
&\sbeam&x&&99.68&658.55&6.95&16043\\%
\hline%
\multirow{8}{*}{\jigsawm}&\sgrad& &\multirow{8}{*}{65.46}&97.62&38.45&8.36&1066\\%
&\sgrad&x&&97.72&29.78&7.56&1151\\%
&\sdel& &&98.71&57.6&7.54&1112\\%
&\sdel&x&&98.63&49.93&6.99&1168\\%
&\sunk&x&&98.75&49.38&6.96&1141\\%
&\sws&x&&98.51&419.58&6.91&12849\\%
&\sgen&x&&88.91&876.81&8.82&26137\\%
&\sbeam&x&&99.54&756.05&7.19&21290\\%
\hline%
\multirow{8}{*}{\tweetm}&\sgrad& &\multirow{8}{*}{96.62}&67.16&63.51&24.13&554\\%
&\sgrad&x&&67.56&49.67&24.08&581\\%
&\sdel& &&71.46&58.78&18.96&412\\%
&\sdel&x&&71.46&48.17&19.37&425\\%
&\sunk&x&&72.23&48.04&19.43&425\\%
&\sws&x&&74.71&178.91&18.81&1467\\%
&\sgen&x&&80.49&1025.66&21.86&6560\\%
&\sbeam&x&&90.07&442.18&18.76&2938\\%
\hline%
\end{tabular}

}
\caption{Effect of with or without part-of-speech constraints when combining with different search strategies on attack performance. The
Search column identifies type of search method used. The POS column identifies if part-of-speech matching constraint is used. 
\label{table:search_full}}
\end{table*}

\begin{table*}[th]
\centering
\scalebox{0.87}{
\begin{tabular}{c|cc|>{\raggedleft\arraybackslash}m{1.8cm}>{\raggedleft\arraybackslash}m{1.8cm}>{\raggedleft\arraybackslash}m{1.8cm}>{\raggedleft\arraybackslash}m{1.6cm}}%
\hline%
Task&Training&Search&Attack Success &Average Number &Average Perturbed &Time (s)\\%
 & & & Rate& of Queries& Word&\\%
\hline%
\multirow{2}{*}{\jigsawb}&\noat&\sunk&99.27&47.62&6.76&939\\%
&\satunk&\sunk&60.28&103.54&13.03&3546\\%
\hline%
\multirow{2}{*}{\jigsawm}&\noat&\sunk&98.75&49.38&6.96&1141\\%
&\satunk&\sunk&71.7&91.08&11.84&3106\\%
\hline%
\multirow{6}{*}{\tweetm}&\noat&\sdel&71.46&48.17&19.37&425\\%
&\noat&\sunk&72.23&48.04&19.43&425\\%
&\satdel&\sdel&21.97&69.68&28.83&598\\%
&\satunk&\sdel&11.17&74.71&33.07&641\\%
&\eat&\sdel&8.28&75.71&27.89&672\\%
&\eat&\sunk&6.08&77.91&33.24&681\\%
\hline%
\end{tabular}

}
\caption{Effect of adversarial training on attack performance on three tasks. 
When attacking, \tae uses the \tglove with $N=20$ plus character transformations; 
constraints with POS; and search with two different greedy methods.  
\label{table:atfull}}
\end{table*}

\begin{table*}[th]
\centering

\begin{tabular}{c|c|rrrr}%
\hline%
Task&Training&AUC&AP&F1&Recall\\%
\hline%
\multirow{2}{*}{\jigsawb}&\noat&0.971&0.749&0.823&0.794\\%
&\satunk&0.971&0.733&0.823&0.811\\%
\hline%
\multirow{2}{*}{\jigsawm}&\noat&0.984&0.636&0.587&0.515\\%
&\satunk&0.984&0.633&0.594&0.535\\%
\hline%
\multirow{4}{*}{\tweetm}&\noat&0.935&0.786&0.730&0.710\\%
&\satdel&0.936&0.792&0.740&0.719\\%
&\satunk&0.938&0.785&0.738&0.723\\%
&\eat&0.932&0.778&0.685&0.641\\%
\hline%
\hline%
\end{tabular}

\caption{Effect of adversarial training on model performance. Macro-average metrics are reported.
\label{table:eval:full}}
\end{table*}

\clearpage
\newpage
\section{Extra on \tae Extensions}
\label{sec:taee}

\subsection{A Suite of Attack Recipes to Extend \tae}

\citet{morris2020reevaluating} splits each text adversarial attack into four parts: goal function, transformation, search strategy and constraints. 
With this modular design, new NLP attacks frequently consist of just  or two new components and often re-use remaining components from past work. We follow this design process, using our newly proposed \tae goal functions to pair with popular choices of other three components from the literature to get a set of \taee recipes.

We select five popular attack recipes from the literature including \deepbugO, \textbuggerO, \atotO, \pwwsO and \textfoolerO that were popular recipes proposed to attack general language classifiers. By swapping these recipes' goal function with the three goal function we propose in \eref{eq:btae} and \eref{eq:mtae}, we construct 15 new \taee attack recipes as shown in Table \ref{table:categorized-attacks}. 
This table categorizes different \taee attack recipes, based on their goal functions, constraints, transformations and search methods. 

\deepbug  and \textbugger are adapted from two SOTA attack recipes \deepbugO \cite{Gao2018BlackBoxGO}  and \textbuggerO \cite{Li2019TextBuggerGA}. They generate \taee  via character manipulations.  These  attacks perform  character insertion, deletion, neighboring swap and replacements to change a word into one that a target toxicity detection model doesn't recognize. These character changes are designed to generate character sequences that a human reader could easily correct into those original words. Language semantics are preserved since human readers can correct the misspellings.

We then select three other popular attacks \atotO, \pwwsO and \textfoolerO \cite{a2t21,pwws-ren-etal-2019-generating, Jin2019TextFooler} to \taee \atot, \pwws and \textfooler attacks. These attacks generate adversarial examples via replacing words from the input with synonyms. These attacks aim to create examples that preserve semantics, grammaticality, and non-suspicion. They  vary regarding the word transformation strategies they use (see Table~\ref{table:categorized-attacks} for details).

All \taee attack recipes use greedy based word importance ranking (Greedy-WIR) strategy to search and determine what words to manipulate (with character changes or with  synonym replacement).

Lastly, these \taee recipes also have difference in what languages constraints they employ to limit the transformations, for instance, \atot puts limit on the number of words to perturb. \textbugger uses universal sentence encoding (USE) similarity as a constraint.

\begin{table*}[h!]
    \centering
    \scalebox{0.79}{
    \begin{tabular}{|p{3.5cm}|p{4cm}|p{6cm}|p{3cm}|}
         \hline
    
    \textbf{Attack Recipe} 
    & \textbf{Constraints} 
    & \textbf{Transformation} 
    & \textbf{Search Method}\\ %
  \hline

     \atot \newline \newline (revised from \cite{a2t21})
     &  Sentence-transformers/all-MiniLM-L6-v2 sentence encoding cosine similarity $> 0.9$ $^\dagger$, Part-of-speech match, Ratio of number of words modified $<0.1$ 
     & Word Synonym Replacement. Swap words with their 20 nearest neighbors in the counter-fitted GLOVE word embedding space or optionally with those predicted by BERT MLM model. 
     & Greedy-WIR (gradient-based)   \\ \hline
    
    \textfooler \newline  (revised from \cite{Jin2019TextFooler}  
    & Word embedding cosine similarity $>0.5$, Part-of-speech match, USE sentence encoding angular similarity $ > 0.84$ 
     & Word Synonym Replacement. Swap words with their 50 nearest neighbors in the counter-fitted GLOVE word embedding space.  
     & Greedy-WIR (deletion-based)
    \\ \hline 
    
    \pwws \newline  (revised from \cite{pwws-ren-etal-2019-generating})  & No special constraints
     & Word Synonym Replacement. Swap words with synonyms from WordNet.  
     & Greedy-WIR (saliency)    \\ \hline

     \deepbug \newline  (revised from \cite{Gao2018BlackBoxGO})
     & Levenshtein edit distance $< 30$
     & \{Random Character Insertion, Random Character Deletion, Random Character Swap, Random Character Replacement\}*
     & Greedy-WIR (deletion-based)      \\ \hline 

     \textbugger  \newline (revised from \cite{Li2019TextBuggerGA})
     & USE sentence encoding cosine similarity $> 0.8$
     & \{Random Character Insertion, Random Character Deletion, Neighboring Character Swap, Character Substitution by Homoglyph $^\ddagger$, Word Synonym Replacement with 5 nearest neighbors in the counter-fitted GLOVE word embedding space\}*   
     &  Greedy-WIR (deletion-based) 
     \\ \hline

    \end{tabular}

    }
    \caption{\taee Attack recipes categorized regarding: search method, transformation, goal function, constraints. All attack recipes include an additional constraint which disallows the replacement of stopwords. Greedy search with Word Importance Ranking (Greedy-WIR). 
    \newline
\footnotesize{   \textit{$\dagger$ indicates new features or enhancements we added to extend the capability of the original attack recipes proposed in the original papers }    
    \newline 
    \textit{* indicates a combination of multiple transformations}. 
    \hspace{3mm}
    \textit{$\ddagger$ substitute a character with visually similar ones or those adjacent on keyboard. For example, substitute "1" with "l". }}
    }
    \label{table:categorized-attacks}
\end{table*}

\tref{table:examples} presents a few generated \taee examples from attacking base models using five \taee attacks.

\begin{table*}[h!]
\centering
\scalebox{0.72}{

\begin{tabular}{|>{\raggedright\arraybackslash}m{3.2cm}>{\raggedright\arraybackslash}m{3.2cm}>{\raggedright\arraybackslash}m{3.2cm}>{\raggedright\arraybackslash}m{3.2cm}>{\raggedright\arraybackslash}m{3.2cm}>{\raggedright\arraybackslash}m{3.2cm}|}%
\hline%
\textbf{Seed Text}&\textbf{\atot}&\textbf{\textfooler}&\textbf{\pwws}&\textbf{\deepbug}&\textbf{\textbugger}\\%
\hline%

\hline%
My neighbors hate me cuz i smoke \textbf{weed} and blast the most ghetto music f...&My neighbors hate me cuz i smoke \textbf{weeds} and blast the most ghetto music fr...&My neighbors hate me cuz i smoke \textbf{pot} and blast the most ghetto music from...&My neighbors hate me cuz i smoke \textbf{smoke} and blast the most ghetto music fr...&My neighbors hate me cuz i smoke \textbf{wee} and blast the most ghetto music from...&My neighbors hate me cuz i smoke \textbf{weеd} and blast the most ghetto music fro...\\%
\hline%
thank god FOB didnt come back doing some \textbf{shitty} arena tour. \#iwaited5ye...&thank god FOB didnt come back doing some \textbf{crappy} arena tour. \#iwaited5year...&thank god FOB didnt come back doing some \textbf{foul} arena tour. \#iwaited5yearsf...&thank god FOB didnt come back doing some \textbf{stinky} arena tour. \#iwaited5year...&thank god FOB didnt come back doing some \textbf{sitty} arena tour. \#iwaited5years...&thank god FOB didnt come back doing some \textbf{crappy} arena tour. \#iwaited5year...\\%
\hline%
Fluffy is a \textbf{retard} homeboy is calling a baby rattlesnake friend and kis...&Fluffy is a \textbf{brainless} homeboy is calling a baby rattlesnake friend and ki...&Fluffy is a \textbf{moron} homeboy is calling a baby rattlesnake friend and kissin...&Fluffy is a \textbf{changeling} homeboy is calling a baby rattlesnake friend and k...&Fluffy is a \textbf{Getard} homeboy is calling a baby rattlesnake friend and kissi...&Fluffy is a \textbf{retarded} homeboy is calling a baby rattlesnake friend and kis...\\%
\hline%
Typically hateful, anti{-}Christian, mentally ill and ugly \textbf{dyke} trash pig ...&Typically hateful, anti{-}Christian, mentally ill and ugly \textbf{levee} trash pig ...&Typically hateful, anti{-}Christian, mentally ill and ugly \textbf{embankment} trash...&Typically hateful, anti{-}Christian, mentally ill and ugly \textbf{dike} trash pig c...&Typically hateful, anti{-}Christian, mentally ill and ugly \textbf{dmyke} trash pig ...&Typically hateful, anti{-}Christian, mentally ill and ugly \textbf{dike} trash pig c...\\%
\hline%
How ghetto \textbf{girls} pronounce there names... I could cry right now oh my g...&How ghetto \textbf{females} pronounce there names... I could cry right now oh my g...&How ghetto \textbf{females} pronounce there names... I could cry right now oh my g...&How ghetto \textbf{missy} pronounce there names... I could cry right now oh my god...&How ghetto \textbf{girsl} pronounce there names... I could cry right now oh my god...&How ghetto \textbf{gilrs} pronounce there names... I could cry right now oh my god...\\%
\hline%
More \textbf{queer} lady Templars. That's my point.&More \textbf{homo} lady Templars. That's my point.&More \textbf{fairy} lady Templars. That's my point.&More \textbf{curious} lady Templars. That's my point.&More \textbf{qIeer} lady Templars. That's my point.&More \textbf{homo} lady Templars. That's my point.\\%
\hline%

\end{tabular}
}
\caption{Selected Toxic Adversarial Examples.  We show adversarial examples generated by attacking base model  \tweetm. To conserve space, we only show results from Offensive Tweet that contain much shorter messages than Jigsaw.  }
\label{table:examples}
\end{table*}

\end{document}